\documentclass[11pt]{article}

\usepackage[]{ACL2023}

\usepackage{times}
\usepackage{latexsym}

\usepackage[T1]{fontenc}

\usepackage[utf8]{inputenc}

\usepackage{microtype}

\usepackage{booktabs}
\usepackage{multirow}
\usepackage{stfloats}
\usepackage{pifont}
\newcommand{\cmark}{\ding{51}}%
\newcommand{\xmark}{\ding{55}}%
\usepackage{xcolor,colortbl}
\usepackage{stfloats}
\usepackage{float}
\usepackage{color}
\definecolor{dkgreen}{rgb}{0,0.6,0}
\definecolor{gray}{rgb}{0.5,0.5,0.5}
\definecolor{mauve}{rgb}{0.58,0,0.82}
\usepackage{listings}
\usepackage{graphicx}
\usepackage{booktabs}
\usepackage{CJKutf8}
\usepackage{inconsolata}
\title{YATO: Yet Another deep learning based Text analysis Open toolkit}

\author{Zeqiang Wang$\phantom{}^{1}$\Thanks{\ Equal contribution.}\textnormal{,}\ Yile Wang$\phantom{}^{2*}$\textnormal{,}\ Jiageng Wu$\phantom{}^{1}$\textnormal{,}\ Zhiyang Teng$\phantom{}^{3}$\textnormal{,}\ Jie Yang$\phantom{}^{1}$\Thanks{\ Corresponding author.} \\
 \textsuperscript{\rm 1}Zhejiang University \\
 \textsuperscript{\rm 2}Institute for AI Industry Research (AIR), Tsinghua University \\
 \textsuperscript{\rm 3}Nanyang Technological University \\
  \textit{zeqiangwang.medicalai@outlook.com} \ \ \  
  \textit{wangyile@air.tsinghua.edu.cn} \\ 
  \textit{jiagengwu@zju.edu.cn}\ \ \ 
\textit{zhiyang.teng@ntu.edu.sg} \ \ \ 
  \textit{jieynlp@gmail.com}
}

\begin{document}
\maketitle
\begin{abstract}
We introduce YATO, an open-source, easy-to-use toolkit for text analysis with deep learning. Different from existing heavily engineered toolkits and platforms, YATO is lightweight and user-friendly for researchers from cross-disciplinary areas. Designed in a hierarchical structure, YATO supports free combinations of three types of widely used features including 1) traditional neural networks (CNN, RNN, \textit{etc}.); 2) pre-trained language models (BERT, RoBERTa, ELECTRA, \textit{etc}.); and 3) user-customized neural features via a simple configurable file. Benefiting from the advantages of flexibility and ease of use, YATO can facilitate fast reproduction and refinement of state-of-the-art NLP models, and promote the cross-disciplinary applications of NLP techniques. The code, examples, and documentation are publicly available at \texttt{\url{https://github.com/jiesutd/YATO}}. A demo video is also available at \texttt{\url{https://www.youtube.com/playlist?list=PLJ0mhzMcRuDUlTkzBfAftOqiJRxYTTjXH}}.
\end{abstract}

\section{Introduction}
Large language models (LLMs) such as GPT-3~\cite{gpt3}, ChatGPT~\cite{chatgpt}, and LLaMA~\cite{llama1,llama2} have gained significant progress in natural language processing (NLP), showing strong abilities to understand text and competitive performance across various NLP tasks. However, these models are either close-source or difficult to fine-tune due to the high computational costs, which makes them inconvenient for academic research or practical implementation.

Alternatively, traditional neural models, such as recurrent neural networks (RNN, \citealp{lstm}), convolutional neural networks (CNN, \citealp{cnn}), and pre-trained language models (PLMs, \citealp{bert,roberta,electra}) have been widely studied and utilized for text understanding. These models benefit from large-scale training data and can be quickly fine-tuned toward specific usages. Recent works also show they can offer useful guidance to LLMs~\cite{xu2023small}. Therefore, small open-source deep learning models are important in current NLP systems, especially in computation and data resource-limited scenarios.

However, due to the complexity of the deep learning model architecture, it is challenging to implement methods or reproduce results from the literature. The different implementations of these models can lead to unfair comparisons or misleading results. Most existing frameworks were designed for professional developers, which brings additional obstacles for less experienced users, especially for researchers with less or no artificial intelligence (AI) background~\cite{zacharias2018survey,zhang2020lattice,johnson2021classical}. 
In addition, these frameworks seldom support user-defined features required for various domain applications (\textit{e.g.}, in medical named entity recognition, customized lexicons can be supplemented as external features, such that additional labels are tagged as features when a word occurs in the lexicon). For non-expert, cross-domain users, customizing models via source code with additional features is complex. To promote interdisciplinary applications of cutting-edge NLP techniques, it is necessary to build a flexible, user-friendly, and effective text representation framework that supports a wide range of deep learning architectures and customized domain features.


There exist several text analysis toolkits in the NLP community. CoreNLP~\citep{corenlp} and spaCy~\citep{spacy2} offer pipelines for many traditional NLP tasks, while the performance is sometimes under-optimal due to the use of less powerful models. AllenNLP~\citep{allennlp} and flairNLP~\citep{flair}  utilize pre-trained models while they do not support user-defined features. FairSeq~\citep{fairseq} is designed for sequence-to-sequence tasks like machine translation and document summarization. Transformers~\citep{transformers} offers implementation for various tasks by using state-of-the-art models across different modalities, while it is heavily engineered. PaddleNLP~\cite{=paddlenlp} and EasyNLP~\cite{easynlp} are specifically designed for industrial application and commercial usage, which are not lightweight for research purposes. The above toolkits are mostly developed for professional AI researchers or engineers, where heavy coding effort is necessary during model development and deployment. The learning curve is steep to fully leverage these toolkits for cross-disciplinary researchers (\textit{e.g.}, medical, financial) who need to build models with lightweight code.

This paper presents a toolkit, \textbf{YATO} (\textbf{Y}et \textbf{A}nother deep learning based \textbf{T}ext analysis \textbf{O}pen toolkit), for researchers looking for a convenient way of building state-of-the-art models for two most popular types of NLP tasks: sequence labeling (\textit{e.g.}, Part-of-Speech tagging, named entity recognition) and sequence classification (\textit{e.g.}, sentiment analysis, document classification). \textsc{YATO} is built on NCRF++ ~\citep{ncrfpp}, a popular neural sequence labeling toolkit with over 250+ citations from research papers, 1,900+ stars and 120+ merged pull requests on GitHub as of Oct. 2023. NCRF++ has been utilized in many cross-disciplinary research projects, including medical \cite{yang2020development} and finance \cite{wan2021sentiment}. YATO retains its strengths, integrates advanced pre-trained language models, and adds capabilities for sequence classification and data visualization.

\begin{table*}[t!]
	\centering
 \small
 \scalebox{0.90}{
	\begin{tabular}{cccccccc}
	    \hline
       \multirow{2}*{\textbf{System}}&\multirow{2}*{\textbf{Lightweight}} &\multirow{2}*{\textbf{PLM}}& \textbf{Neural} &\textbf{User-Defined} &\textbf{Configurable} & \textbf{SOTA}&\multirow{2}*{\textbf{Reference}}  \\
       & &  &\textbf{Models}&\textbf{Features}&\textbf{w/o Coding} & \textbf{Performance}&  \\
    	 \hline
            
    	CoreNLP&\cmark&\xmark &\xmark& \xmark& \xmark&\xmark&\cite{corenlp}\\
    	spaCy &\cmark&\cmark &\cmark& \xmark& \xmark&\xmark&\cite{spacy2}\\
    	 AllenNLP&\cmark&\cmark &\cmark&\xmark&\xmark&  \cmark&\cite{allennlp} \\
    	FlairNLP&\cmark&\cmark &\cmark&\xmark&\xmark& \cmark&\cite{flair} \\
           NCRF++&\cmark&\xmark &\cmark& \cmark& \cmark&\cmark&\cite{ncrfpp} \\
    	  FairSeq &\xmark&\cmark &\cmark&\xmark&\xmark&  \cmark&\cite{fairseq} \\
    	   Transformers&\xmark&\cmark&\cmark&\xmark&\xmark& \cmark&\cite{transformers} \\
    	   PaddleNLP&\xmark&\cmark&\cmark & \xmark& \xmark&\cmark&\cite{=paddlenlp} \\
    	   EasyNLP&\xmark&\cmark&\cmark &\xmark &\xmark &\cmark  &\cite{easynlp} \\
    	   \hline
    	  \textsc{\textbf{YATO}}&\cmark&\cmark  &\cmark&\cmark&\cmark& \cmark& \textbf{This paper}  \\
    	\hline
	\end{tabular}}
	\caption{Comparison between existing popular text analysis libraries and our proposed \textsc{YATO}.} 
	\label{table:comparison}
\end{table*}




\section{Highlights of YATO}
Table~\ref{table:comparison} lists the comparison of YATO and popular existing text analysis libraries. The highlights of YATO include:


\noindent{ \bf $\bullet$ Lightweight.} \textsc{YATO} focuses on two fundamental while popular NLP tasks: sequence labeling and sequence classification, covering many downstream applications such as information extraction, sentiment analysis, text classification, \textit{etc}. Different from the heavily engineered libraries, YATO is concise and lightweight with less library dependence. It can be fast developed and deployed in various environments, making it a  user-friendly toolkit for less experienced users.

\begin{figure*}[t!]
    \centering
    \resizebox{0.8\linewidth}{!}{
        \includegraphics{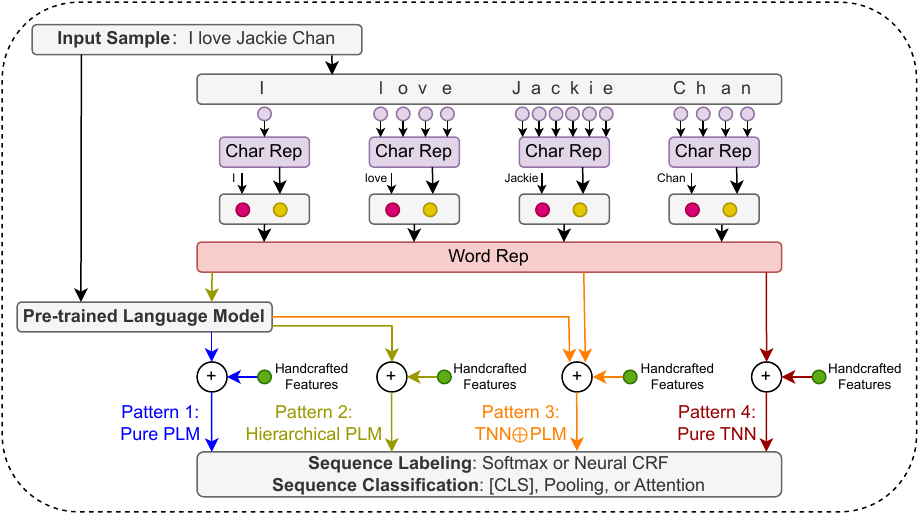}
    }

    \caption{The overall architecture of YATO. The purple, red, yellow, and green circles represent character embeddings, word embeddings, character sequence representation, and custom handcrafted features, respectively.}
    \label{fig:overall_framework}

\end{figure*}

\noindent{ \bf $\bullet$ Flexible.}  Most of the existing libraries do not support the combination of various neural features. By using YATO, users can customize their models through free combinations of various neural models, including traditional neural networks (CNN, RNN) and state-of-the-art PLMs, as well as handcrafted features for domain adaptation. YATO also supports various inference layers, including attention pooling, softmax, conditional random field (CRF), and \textit{nbest} decoding. 

\noindent{ \bf $\bullet$ Configurable.} To minimize the effort of coding, all the model developments on YATO can be easily conducted by editing the configuration file. YATO will load the configuration file and construct the deep learning models following the configurations. 

\noindent{ \bf $\bullet$ Easy to Use.} \textsc{YATO} is built based on PyTorch\footnote{\url{https://pytorch.org/}.} and has been released on PyPI\footnote{\url{https://pypi.org/}.}, the installation can be done through \textit{pip install ylab-yato}. For non-AI users, editing a configuration file to build deep learning models is simple and intuitive. For AI users, \textsc{YATO} provides various modularized functions for professional development. 

\noindent{ \bf $\bullet$ High Performance.} In extensive experiments on sequence labeling and classification tasks, YATO proves that it can achieve state-of-the-art performance on most tasks and datasets. YATO offers flexibility in terms of hardware resources, supporting both GPU and CPU for training and inference processes. It provides the ability to specify the desired device configuration, facilitating efficient utilization of multiple GPUs on a single server.

\noindent{ \bf $\bullet$ Visualization.} \textsc{YATO} offers the interface for visualizing text attention, which can help users further interpret and analyze the results.

\section{Architecture of YATO}
YATO is designed \textit{hierarchically} to support flexible combinations of character-level features, word-level features, and pre-trained language models\footnote{It supports all available pre-trained models from \url{https://huggingface.co/models}.}, as well as handcrafted features. As illustrated in  Figure~\ref{fig:overall_framework}, YATO supports four patterns to represent text as embeddings and with flexible choices on adding handcrafted features and inference layers.

\subsection{Text Representations}
\noindent{\textbf{Pure Pre-trained Language Model}.} YATO enables the initialization of parameters with pre-trained language models such as BERT~\cite{bert}, RoBERTa~\cite{roberta}, and ELECTRA~\cite{electra}, and fine-tunes them on training data. Leveraging the rich knowledge inside the PLM, they have demonstrated strong performance on downstream tasks. To better leverage the models with domain-specific knowledge, YATO also supports pre-trained models designed for specific tasks, such as SciBERT~\cite{Beltagy2019SciBERT},  BioBERT~\cite{lee2020biobert} and others.

\noindent{\textbf{Hierarchical Pre-trained Language Model}.} The hierarchical pre-trained language model in YATO differs from the conventional notion of hierarchy, which typically describes relationships between word, sentence, and document structures. Instead, it signifies the ordinal relation between the traditional neural network and the pre-trained language model. Specifically, YATO supports using both word sequence features and the pre-trained language model representations in a hierarchical way, where the word and character features can be explicitly encoded in advance and used as the input for the pre-trained language model.

\noindent{\textbf{Traditional Neural Network (TNN) \& Pre-trained Language Model}.} In contrast to the hierarchical combination, we can use the word sequence features directly before the final prediction layer, combined with the representation brought from the PLMs. Such a feature-based approach is also used in ELMo~\cite{elmo} and BERT~\cite{bert}, which shows close performance while does not require fine-tuning the pre-trained models.

\noindent{\textbf{Pure Traditional Neural Network}.} Besides using the transformer-based pre-trained models, we also support traditional neural models such as RNN, CNN, and BiLSTM. Compared with Transformer, these models usually have fewer parameters and are also shown effective for sequence modeling~\cite{ma-hovy-2016-end,lample-etal-2016-neural,yang-etal-2018-design}, especially when the training data is limited.

\subsection{Handcrafted Features and Inference}
\noindent{\textbf{Handcrafted Features }.} YATO provides feature embedding modules to encode any handcrafted features and the corresponding feature embeddings can be concatenated to the text representation on any patterns. Users can add feature embeddings by formatting the text following YATO instructions without any coding effort.  

\noindent {\textbf{Inference Layers.}} The encoded text embeddings from the text representation structure are fed into a softmax or CRF ~\cite{crf} layer for sequence labeling tasks, and YATO also supports \textit{nbest} decoding to generate more candidate label sequences with probabilities for further optimization. For sequence classification tasks, a classifier head over [CLS] representation, or pooling/attention layer on all word representations is used, and a visualization tool is also available in YATO to visualize the word importance in the attention layer.

\section{The Overall Workflow and Usage}


The overall workflow of YATO comprises three primary components to facilitate solving downstream tasks: configuration preparation, model training, and model decoding. Users can leverage this modular pipeline to tackle a wide range of applications.




\begin{table}[t!]
    \centering
    \resizebox{\linewidth}{!}{
    \begin{tabular}{l}
    \toprule
    \rowcolor[gray]{0.95} {\textbf{Configuration File}} \\
    \\
    \#\textbf{Dataloader}\\
    \texttt{train\_dir=The path of train dataset}\\
    \texttt{dev\_dir=The path of development dataset}\\
    ...\\
    \#\textbf{Model}\\ 
    \texttt{use\_crf=True/False}\\
    \texttt{use\_char=True/False}\\
    \texttt{char\_seq\_feature=GRU/LSTM/CNN/False}\\
     \texttt{word\_seq\_feature=GRU/LSTM/CNN/FeedFowrd/False}\\
 \texttt{low\_level\_transformer=pretrain language model}\\
    \texttt{high\_level\_transformer=PLM from huggingface}\\
    \texttt{bilstm=True/False}\\
    ...\\
        \#\textbf{Handcrafted Features}\\
  \texttt{feature=[POS] emb\_size=20 emb\_dir=your POS embedding}\\
\texttt{feature=[Cap] emb\_size=20 emb\_dir=your Cap embedding}\\
...\\
    \#\textbf{Hyperparameters }\\
    \texttt{sentence\_classification=True/False}\\
    \texttt{status=train/decode}\\
    \texttt{iteration=epoch number}\\         
    \texttt{batch\_size=batch size}\\ 
    \texttt{optimizer=SGD/Adagrad/adadelta/rmsprop/adam/adamw}\\
    \texttt{learning\_rate=learning rate}\\
    ...\\
    \#\textbf{Prediction}\\
    \texttt{raw\_dir=The path of decode file}\\
    \texttt{decode\_dir=The path of the decode result file}\\
        \texttt{nbest=0 (for labeling)/1 (for classification)}\\
    ...\\
    \bottomrule
    \end{tabular}}
        \caption{A sample of a configuration file.}
    \label{tab:config_prepare}
\end{table}

\noindent{\textbf{Configuration Preparation.}} Users can specify the dataset, model, optimizer, and decoding through the configuration file, as shown in Table~\ref{tab:config_prepare}. In particular, different patterns can be customized by setting the values of ``high\_level\_transformer'', ``low\_level\_transformer'', ``bilstm'', and different word sequence representations can be easily designed through specifying ``char\_seq\_feature'' and ``word\_seq\_feature''. Handcrafted features can be added through ``feature''.


\noindent{\textbf{Model Training.}}
YATO enables efficient training of high-performance models for sequence labeling and sentence classification with minimal code and configuration file specification. For example, users can train a competitive named entity recognition or text classification model using a few lines of code and a configuration file. The simplicity of the YATO interface allows rapid prototyping and experimentation for these fundamental NLP tasks. The framework was designed in batch computation which can fully utilize the power of GPUs.


\noindent{\textbf{Model Decoding.}} Similar to model training, simple file configuration can be used to enable YATO. Besides the greedy decoding, \textsc{YATO} also supports \textit{nbest} decoding, \textit{i.e.}, which decodes label sequences with the top n probabilities by using the Viterbi decoding in neural CRF layers. The \textit{nbest} results can serve as important resources for further optimizations, \textit{e.g.}, reranking~\cite{yang2017neural}.

\section{Experiments}

\subsection{Datasets and Main Results}
To evaluate our framework, we evaluated 8 datasets that cover sequence labeling and classification tasks in both English and Chinese, including  
named entity recognition (NER) on CoNLL2003 \citep{conll2003ner}, OntoNotes~\citep{hovy-etal-2006-ontonotes} and MSRA~\citep{levow-2006-third}; CCG supertagging on CCGBank~\citep{ccgbank}; sentiment analysis on SST2, SST5~\citep{sst5}, and ChnSentiCorp~\citep{TAN20082622}.

\begin{table*}[t!]
  \centering
  \small
   \resizebox{\linewidth}{!}{
\begin{tabular}{ccccccccccc}
    \toprule
    \multirow{2}[2]{*}{\textbf{Model}} & \multicolumn{2}{c}{\textbf{CoNLL 2003}} & \multicolumn{2}{c}{\textbf{OntoNotes 5.0}} & \multicolumn{2}{c}{\textbf{MSRA}} & \multicolumn{2}{c}{\textbf{Ontonotes 4.0}}  & \multicolumn{2}{c}{\textbf{CCG supertagging}} \\
\cmidrule(lr){2-3}\cmidrule(lr){4-5}\cmidrule(lr){6-7}\cmidrule(lr){8-9}\cmidrule(lr){10-11}
& \textbf{YATO} & \textbf{Ref.} & \textbf{YATO} & \textbf{Ref.} & \textbf{YATO} & \textbf{Ref.} & \textbf{YATO} & \textbf{Ref.} & \textbf{YATO} & \textbf{Ref.} \\
    \midrule
    CCNN+WLSTM+CRF & \textbf{91.26}    & 91.11$^a$  & 81.53 & -     & 92.83 & -     & 74.55 & -      & 93.80  & - \\
    BERT-base & 91.61 & 92.4$^b$& 84.68 & 85.54$^e$ & \textbf{95.81} & 94.71$^f$ &  \textbf{80.57}& 79.93$^f$  & \textbf{96.14} & 92.16$^h$  \\
    RoBERTa-base &  \textbf{90.23}& 90.11$^c$ & \textbf{86.28} & 86.2$^c$ & \textbf{96.02} & 95.1$^g$ & \textbf{80.94} & 80.37$^e$& 96.16 & - \\
    ELECTRA-base & \textbf{91.59}  & 91.5$^d$ & 85.25 & 87.6$^d$  & 96.03 & -     & 80.47 & -   & 96.29 & - \\
    \bottomrule
    \end{tabular}%

}

\caption{Results for sequence labeling tasks. \textbf{Bold} represents that YATO's re-produce is more accurate. $^a$\citep{yang-etal-2018-design} $^b$\citep{bert} $^c$\citep{bond2020} $^d$\citep{electra-res-active} $^e$\citep{wang2022k} $^f$\citep{liu-etal-2021-lexicon} $^g$\citep{li2022markbert} $^h$\citep{2020-ccg}}
\label{table:results_labeling}
\end{table*}%

\begin{table*}[t!]
  \centering
    \small
    \begin{tabular}{ccccccc}
    \toprule
    \multirow{2}[2]{*}{\textbf{Model}} & \multicolumn{2}{c}{\textbf{SST2}} & \multicolumn{2}{c}{\textbf{SST5}} &  \multicolumn{2}{c}{\textbf{ChnSentiCorp}}  \\
\cmidrule(lr){2-3}\cmidrule(lr){4-5}\cmidrule(lr){6-7}
& \textbf{YATO} & \textbf{Ref.} & \textbf{YATO} & \textbf{Ref.} & \textbf{YATO} & \textbf{Ref.} \\
    \midrule
    CCNN+WLSTM+CRF & 87.61 &  -     & 43.48 & -      & 88.22 &  -     \\
    BERT-base & \textbf{93.00}$^\dag$ & 92.7$^\dag$$^a$      & \textbf{53.48} & 53.2$^d$      & \textbf{95.96} &    95.3$^g$ \\
    RoBERTa-base &  92.55$^\dag$ & 94.8$^\dag$$^b$       & 51.99 &  56.4$^e$   & \textbf{96.04} & 95.6$^h$    \\
    ELECTRA-base & 94.72$^\dag$ & 95.1$^\dag$$^c$        & \textbf{55.11} & 54.8$^f$     & \textbf{95.96} &   94.5$^g$    \\
    \bottomrule
    \end{tabular}%

\caption{Results of sequence classification tasks. \textbf{Bold} represents that YATO re-produce is more accurate.  \dag\ denotes the results of the dev set. $^a$\citep{bert} $^b$\citep{zaheer2020big} $^c$\citep{electra} $^d$\citep{munikar2019fine} $^e$\citep{sun2020self}  $^f$\citep{xia2022prompting} $^g$\citep{wwmcbert2921} $^h$\citep{xin-etal-2020-deebert}}
\label{table:results_classification}
\end{table*}%

Table~\ref{table:results_labeling} and Table~\ref{table:results_classification} demonstrate that YATO can reproduce both classical and state-of-the-art deep learning models on most sequence labeling and classification tasks. For some results such as BERT on CoNLL, the originally reported 92.4 F1 score by \citet{bert} may not be achieved with current libraries, as discussed in previous literature~\cite{stanislawek-etal-2019-named,gui-etal-2020-uncertainty}. Overall, YATO achieves the best performance on MSRA,  OntoNotes 4.0, CCG supertagging, and ChnSentiCorp. The compatibility and reproducibility across different models and tasks demonstrate that \textsc{YATO} can serve as a platform for reproducing and comparing different methods from classical neural models to state-of-the-art PLMs.

\begin{table}[t]
  \centering
    \small
    \begin{tabular}{lccccc}
    \toprule
    {\textbf{Patterns}}&\textbf{SST5}  & \textbf{CoNLL 2003} \\
    \midrule
    1. Pure PLM&53.48&\textbf{91.61}\\
    \midrule
    2. Hierarchical PLM&53.77&90.52\\
    \midrule
    3. TNN$\oplus$ PLM&\textbf{54.84}&90.47\\
    \midrule
    4. Pure TNN&43.48&91.26\\
    \bottomrule
    \end{tabular}%
    \caption{Performances of different training patterns.}
    \label{tab:pattern}
\end{table}

\subsection{Comparison of Different Patterns}
 Table~\ref{tab:pattern} shows the performance of four different model patterns on both sequence labeling and classification tasks (one dataset for each task). The combination of Hierarchical PLM and TNN$\oplus$PLM (patterns 2 and 3) outperforms pure models (patterns 1 and 4) on SST5. However, pure PLM achieves the best performance on the CoNLL 2003 NER dataset. These results demonstrate that complex models are not always better than simple models, and a flexible framework is necessary for providing various model candidates.
\begin{table}[t]
  \centering
    \small
    \resizebox{.48\textwidth}{!}{ 
\begin{tabular}{ccc}
    \toprule
          & \textbf{NCBI-disease} & \textbf{Yidu-s4k} \\
    \midrule
    Pure PLM & 84.23 & 82.73 \\
    \midrule
    \multicolumn{1}{c}{\multirow{2}[4]{*}{+Handcrafted Features}} & \textbf{84.85} (Cap) & 82.98 (Lexicon2) \\
\cmidrule{2-3}          & 84.78 (Lexicon1) & \textbf{83.46} (Lexicon3) \\
    \bottomrule
        \end{tabular}}

    \caption{Handcraft features. Lexicon1 (English medical glossary), Lexicon2 (medical word list from THUOCL), and Lexicon3 (Chinese medical glossary).}
    \label{tab:handcraft}
\end{table}

\subsection{Results by Using Handcraft Features}
To demonstrate the effectiveness of encoding handcraft features in domain application, Table~\ref{tab:handcraft} shows the comparison results on two medical NER tasks, the NCBI-disease~\cite{dougan2014ncbi} for English and Yidu-S4K~\cite{yidu4k} for Chinese. Experiments on NCBI-disease apply two types of features, capitalization and English medical lexicon from the Chinese-English mapping medical glossary\footnote{\url{http://medtop10.com}\label{medtop} }. Experiments on Yidu-s4k dataset employ two medical lexicons as handcrafted features: the medical glossary of THUOCL~\cite{han2016thuocl} (THU Open Chinese Lexicon) and the same medical glossary sourced from the web\textsuperscript{\ref {medtop}} while in Chinese version. Results show that handcraft features can improve the model performance in the medical domain.

\begin{figure*}[htbp]
    \centering
    \resizebox{0.98\linewidth}{!}{
        \includegraphics{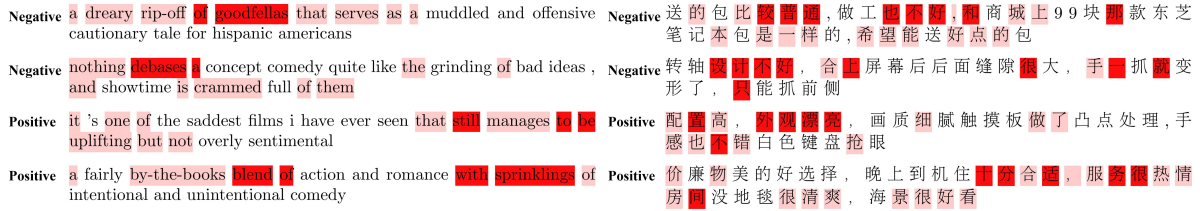}
    }
    \caption{Visualization of attention weights. Different degrees of background color reflect the distributions of attention in words or characters.}
    \label{fig:vis}

\end{figure*} 

\begin{table}[t]
  \centering 
\resizebox{.48\textwidth}{!}{ 
\begin{tabular}{ccc}
    \toprule
    \textbf{Tasks (Metric)}      & \textbf{Transformers} & \textbf{YATO} \\
    \midrule
    MRPC (Accuracy) & 84.07 &  \textbf{85.78}\\
    QQP (Accuracy) &  90.71 &  \textbf{90.81}\\
    MNLI (Matched Accuracy) &  83.91 &  \textbf{84.21}\\
    QNLI (Accuracy) &  \textbf{90.66} &  90.28\\
    SST2 (Accuracy) & 92.32  &  \textbf{93.00}\\
    RTE (Accuracy) &  \textbf{65.70} &  63.90\\
    CoLA (Correlation Coefficient) &  56.53 &   \textbf{58.00}\\
    \midrule
    \textbf{Average} &  80.55 & \textbf{80.85} \\
   
    \bottomrule
    \end{tabular}
}
    \caption{Results of fine-tuning BERT-base-uncased model on GLUE benchmark.}
    \label{tab:glue}
\end{table}

\subsection{Comparison with Transformers}
The aforementioned results show that we can achieve the reported values across various tasks by using \textsc{YATO}. We further use tasks from GLUE benchmark~\cite{wang2018glue} and compare with the results by using Huggingface Transformers~\cite{transformers}, which is one of the most popular libraries. Table~\ref{tab:glue} shows the results by using BERT-base-uncased model, the values of Huggingface Transformers are sourced from the corresponding github page\footnote{\url{https://github.com/huggingface/transformers/tree/main/examples/pytorch/text-classification}}. YATO  achieves comparable and overall better performance than that of Huggingface Transformers by using default settings.

\subsection{Visualization of Attention Map}

Beyond performance, \textsc{YATO} provides a visualization tool for taking the list of words and the corresponding weights as input to generate Latex code for visualizing the attention-based result. Figure~\ref{fig:vis} provides visualization examples of attention on sentiment prediction tasks. Words or characters with sentiment polarities can be automatically extracted and highlighted using our YATO module. As shown in this table, words that have a high impact on the sentiment are highlighted. This visualization module can improve the interpretability of deep learning models in our toolkit.

\begin{figure}[t]
    \centering
        \includegraphics[width=0.5\textwidth]{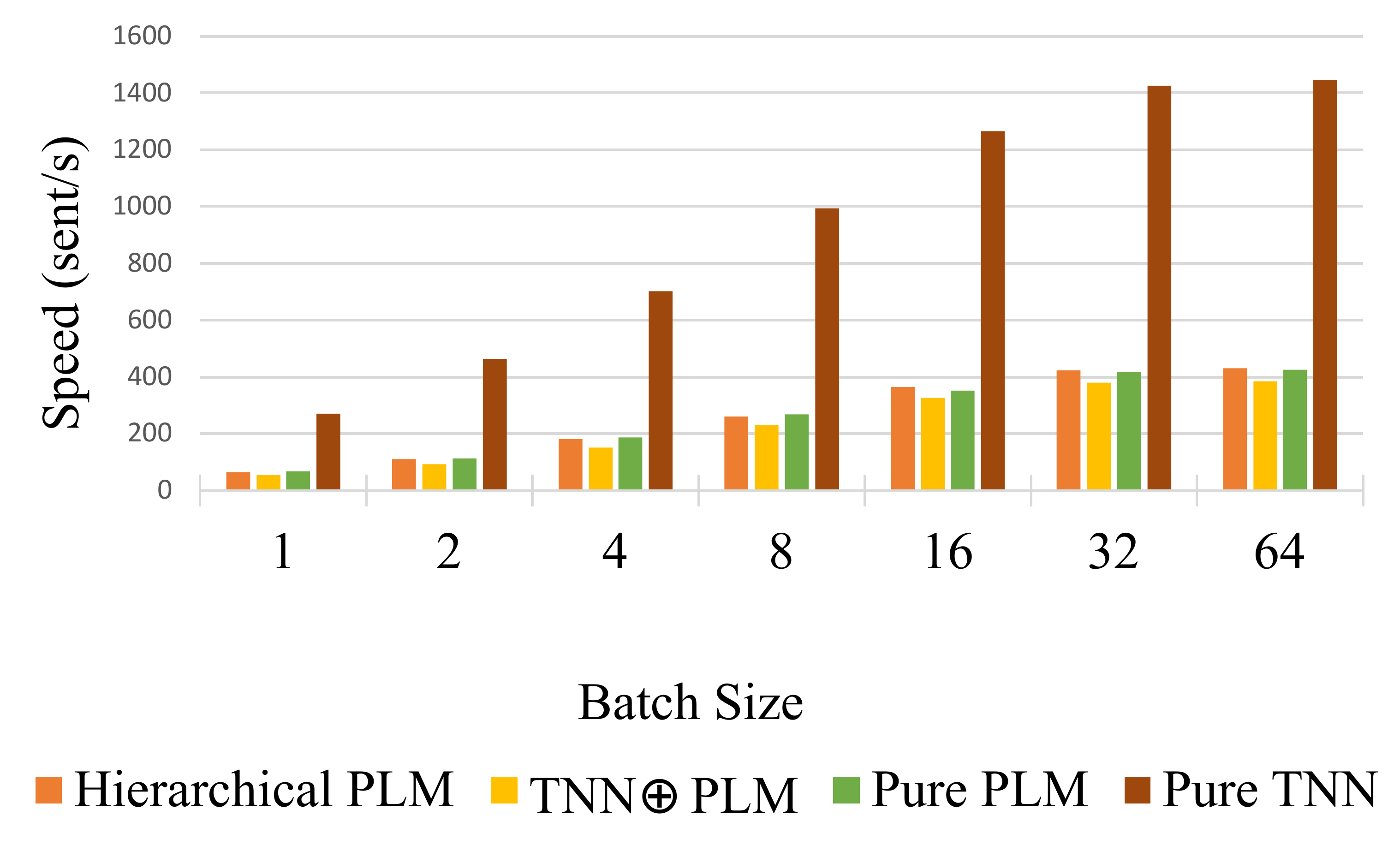}
    \caption{Decoding speed for the four patterns on different batch sizes. Tested on NVIDIA RTX 2080Ti GPU.}
    \label{fig:speed}
\end{figure} 

\subsection{Efficiency Analysis}
YATO is implemented using a fully batch computing approach, making it quite efficient in both model training and decoding. With the help of GPU and large batches, models built on YATO can be decoded efficiently. Figure~\ref{fig:speed} shows the decoding speed of four patterns at different batch sizes. The Pure TNN model has the best inference speed (1400 sentences/s) with a batch size of 64. The decoding speed of Transformer-based models decreases to around 400 sentences/s, showing the trade-off between efficiency and performance. Overall, without using external optimization techniques, YATO has a competitive decoding speed. 

\section{Conclusion}

\textsc{YATO} is an open-source toolkit for text analysis that supports various combinations of state-of-the-art deep learning models and user-customed features, with high flexibility and minimum coding effort. \textsc{YATO} is maintained by core developers from YLab (\texttt{\url{https://ylab.top/}}). It aims to help AI researchers build state-of-the-art NLP models and assist non-AI researchers in conducting cross-disciplinary research with advanced NLP techniques. Given the success of its predecessor, NCRF++, we believe that YATO will greatly promote the applications of NLP in various cross-disciplinary fields and reduce disparities of AI application in these areas. In the future, we plan to integrate advanced LLMs and customize modules that support modeling time series, multimodal features, and specific features for various domains. 

\section*{Limitations}
Our proposed text analysis toolkit mainly focuses on discriminative style tasks, where most of them are treated as token-level or sentence-level classification tasks. Recent studies show that the generative style language models such as GPT~\cite{radford2018improving}, BART~\cite{lewis2019bart}, and T5~\cite{raffel2020exploring} can also show promising zero-shot and few-shot results by adding user-defined prompts or instructions as external inputs, we leave this as our future work.


\bibliographystyle{acl_natbib}
\bibliography{anthology_bak,custom}

\begin{thebibliography}{59}
\expandafter\ifx\csname natexlab\endcsname\relax\def\natexlab#1{#1}\fi

\bibitem[{Akbik et~al.(2019)Akbik, Bergmann, Blythe, Rasul, Schweter, and
  Vollgraf}]{flair}
Alan Akbik, Tanja Bergmann, Duncan Blythe, Kashif Rasul, Stefan Schweter, and
  Roland Vollgraf. 2019.
\newblock Flair: An easy-to-use framework for state-of-the-art nlp.
\newblock In \emph{Proc. of NAACL, (Demonstrations)}.

\bibitem[{Beltagy et~al.(2019)Beltagy, Lo, and Cohan}]{Beltagy2019SciBERT}
Iz~Beltagy, Kyle Lo, and Arman Cohan. 2019.
\newblock \href {http://arxiv.org/abs/arXiv:1903.10676} {Scibert: Pretrained
  language model for scientific text}.
\newblock In \emph{EMNLP}.

\bibitem[{Brown et~al.(2020)Brown, Mann, Ryder, Subbiah, Kaplan, Dhariwal,
  Neelakantan, Shyam, Sastry, Askell, Agarwal, Herbert-Voss, Krueger, Henighan,
  Child, Ramesh, Ziegler, Wu, Winter, Hesse, Chen, Sigler, Litwin, Gray, Chess,
  Clark, Berner, McCandlish, Radford, Sutskever, and Amodei}]{gpt3}
Tom Brown, Benjamin Mann, Nick Ryder, Melanie Subbiah, Jared~D Kaplan, Prafulla
  Dhariwal, Arvind Neelakantan, Pranav Shyam, Girish Sastry, Amanda Askell,
  Sandhini Agarwal, Ariel Herbert-Voss, Gretchen Krueger, Tom Henighan, Rewon
  Child, Aditya Ramesh, Daniel Ziegler, Jeffrey Wu, Clemens Winter, Chris
  Hesse, Mark Chen, Eric Sigler, Mateusz Litwin, Scott Gray, Benjamin Chess,
  Jack Clark, Christopher Berner, Sam McCandlish, Alec Radford, Ilya Sutskever,
  and Dario Amodei. 2020.
\newblock \href
  {https://proceedings.neurips.cc/paper_files/paper/2020/file/1457c0d6bfcb4967418bfb8ac142f64a-Paper.pdf}
  {Language models are few-shot learners}.
\newblock In \emph{Advances in Neural Information Processing Systems},
  volume~33, pages 1877--1901. Curran Associates, Inc.

\bibitem[{CCKS(2019)}]{yidu4k}
CCKS. 2019.
\newblock China conference on knowledge graph and semantic computing 2019 named
  entity recognition of chinese electronic medical record.
\newblock \url{https://www.sigkg.cn/ccks2019/en/}.

\bibitem[{Clark et~al.(2020)Clark, Luong, Le, and Manning}]{electra}
Kevin Clark, Minh-Thang Luong, Quoc~V. Le, and Christopher~D. Manning. 2020.
\newblock {ELECTRA}: Pre-training text encoders as discriminators rather than
  generators.
\newblock In \emph{Proc. of ICLR}.

\bibitem[{Contributors(2021)}]{=paddlenlp}
PaddleNLP Contributors. 2021.
\newblock Paddlenlp: An easy-to-use and high performance nlp library.
\newblock \url{https://github.com/PaddlePaddle/PaddleNLP}.

\bibitem[{Cui et~al.(2021)Cui, Che, Liu, Qin, and Yang}]{wwmcbert2921}
Yiming Cui, Wanxiang Che, Ting Liu, Bing Qin, and Ziqing Yang. 2021.
\newblock Pre-training with whole word masking for {C}hinese {BERT}.
\newblock \emph{IEEE/ACM Trans. Audio, Speech and Lang. Proc.}, 29:3504–3514.

\bibitem[{Devlin et~al.(2019)Devlin, Chang, Lee, and Toutanova}]{bert}
Jacob Devlin, Ming-Wei Chang, Kenton Lee, and Kristina Toutanova. 2019.
\newblock {BERT}: Pre-training of deep bidirectional transformers for language
  understanding.
\newblock In \emph{Proc. of NAACL}.

\bibitem[{Do{\u{g}}an et~al.(2014)Do{\u{g}}an, Leaman, and Lu}]{dougan2014ncbi}
Rezarta~Islamaj Do{\u{g}}an, Robert Leaman, and Zhiyong Lu. 2014.
\newblock Ncbi disease corpus: a resource for disease name recognition and
  concept normalization.
\newblock \emph{Journal of biomedical informatics}, 47:1--10.

\bibitem[{Durrani et~al.(2020)Durrani, Sajjad, Dalvi, and Belinkov}]{2020-ccg}
Nadir Durrani, Hassan Sajjad, Fahim Dalvi, and Yonatan Belinkov. 2020.
\newblock Analyzing individual neurons in pre-trained language models.
\newblock In \emph{Proc. of EMNLP}.

\bibitem[{Gardner et~al.(2017)Gardner, Grus, Neumann, Tafjord, Dasigi, Liu,
  Peters, Schmitz, and Zettlemoyer}]{allennlp}
Matt Gardner, Joel Grus, Mark Neumann, Oyvind Tafjord, Pradeep Dasigi,
  Nelson~F. Liu, Matthew Peters, Michael Schmitz, and Luke~S. Zettlemoyer.
  2017.
\newblock \href {http://arxiv.org/abs/arXiv:1803.07640} {Allennlp: A deep
  semantic natural language processing platform}.

\bibitem[{Gui et~al.(2020)Gui, Ye, Zhang, Li, Fei, Gong, and
  Huang}]{gui-etal-2020-uncertainty}
Tao Gui, Jiacheng Ye, Qi~Zhang, Zhengyan Li, Zichu Fei, Yeyun Gong, and
  Xuanjing Huang. 2020.
\newblock \href {https://doi.org/10.18653/v1/2020.emnlp-main.181}
  {Uncertainty-aware label refinement for sequence labeling}.
\newblock In \emph{Proceedings of the 2020 Conference on Empirical Methods in
  Natural Language Processing (EMNLP)}, pages 2316--2326, Online. Association
  for Computational Linguistics.

\bibitem[{Han et~al.(2016)Han, Zhang, Ma, Tu, Guo, Liu, and
  Sun}]{han2016thuocl}
Shiyi Han, Yuhui Zhang, Yunshan Ma, Cunchao Tu, Zhipeng Guo, Zhiyuan Liu, and
  Maosong Sun. 2016.
\newblock Thuocl: Tsinghua open chinese lexicon.
\newblock \emph{Tsinghua University}.

\bibitem[{Hochreiter and Schmidhuber(1997)}]{lstm}
Sepp Hochreiter and J{\"u}rgen Schmidhuber. 1997.
\newblock Long short-term memory.
\newblock \emph{Neural computation}, 9(8):1735--1780.

\bibitem[{Hockenmaier and Steedman(2007)}]{ccgbank}
Julia Hockenmaier and Mark Steedman. 2007.
\newblock {CCGbank: A Corpus of CCG Derivations and Dependency Structures
  Extracted from the Penn Treebank}.
\newblock \emph{Computational Linguistics}, 33(3):355--396.

\bibitem[{Honnibal and Montani(2017)}]{spacy2}
Matthew Honnibal and Ines Montani. 2017.
\newblock spacy 2: Natural language understanding with bloom embeddings,
  convolutional neural networks and incremental parsing.

\bibitem[{Hovy et~al.(2006)Hovy, Marcus, Palmer, Ramshaw, and
  Weischedel}]{hovy-etal-2006-ontonotes}
Eduard Hovy, Mitchell Marcus, Martha Palmer, Lance Ramshaw, and Ralph
  Weischedel. 2006.
\newblock \href {https://aclanthology.org/N06-2015} {{O}nto{N}otes: The 90{\%}
  solution}.
\newblock In \emph{Proceedings of the Human Language Technology Conference of
  the {NAACL}, Companion Volume: Short Papers}, pages 57--60, New York City,
  USA. Association for Computational Linguistics.

\bibitem[{Johnson et~al.(2021)Johnson, Burns, Stewart, Cook, Besnier, and
  Mattingly}]{johnson2021classical}
Kyle~P Johnson, Patrick~J Burns, John Stewart, Todd Cook, Cl{\'e}ment Besnier,
  and William~JB Mattingly. 2021.
\newblock The classical language toolkit: An nlp framework for pre-modern
  languages.
\newblock In \emph{Proceedings of the 59th annual meeting of the association
  for computational linguistics and the 11th international joint conference on
  natural language processing: System demonstrations}, pages 20--29.

\bibitem[{Lafferty et~al.(2001)Lafferty, McCallum, and Pereira}]{crf}
John~D. Lafferty, Andrew McCallum, and Fernando C.~N. Pereira. 2001.
\newblock Conditional random fields: Probabilistic models for segmenting and
  labeling sequence data.
\newblock In \emph{Proc. of ICML}.

\bibitem[{Lample et~al.(2016)Lample, Ballesteros, Subramanian, Kawakami, and
  Dyer}]{lample-etal-2016-neural}
Guillaume Lample, Miguel Ballesteros, Sandeep Subramanian, Kazuya Kawakami, and
  Chris Dyer. 2016.
\newblock \href {https://doi.org/10.18653/v1/N16-1030} {Neural architectures
  for named entity recognition}.
\newblock In \emph{Proceedings of the 2016 Conference of the North {A}merican
  Chapter of the Association for Computational Linguistics: Human Language
  Technologies}, pages 260--270, San Diego, California. Association for
  Computational Linguistics.

\bibitem[{LeCun et~al.(1989)LeCun, Boser, Denker, Henderson, Howard, Hubbard,
  and Jackel}]{cnn}
Y.~LeCun, B.~Boser, J.~S. Denker, D.~Henderson, R.~E. Howard, W.~Hubbard, and
  L.~D. Jackel. 1989.
\newblock \href {https://doi.org/10.1162/neco.1989.1.4.541} {Backpropagation
  applied to handwritten zip code recognition}.
\newblock \emph{Neural Computation}, 1(4):541--551.

\bibitem[{Lee et~al.(2020)Lee, Yoon, Kim, Kim, Kim, So, and
  Kang}]{lee2020biobert}
Jinhyuk Lee, Wonjin Yoon, Sungdong Kim, Donghyeon Kim, Sunkyu Kim, Chan~Ho So,
  and Jaewoo Kang. 2020.
\newblock Biobert: a pre-trained biomedical language representation model for
  biomedical text mining.
\newblock \emph{Bioinformatics}, 36(4):1234--1240.

\bibitem[{Levow(2006)}]{levow-2006-third}
Gina-Anne Levow. 2006.
\newblock \href {https://aclanthology.org/W06-0115} {The third international
  {C}hinese language processing bakeoff: Word segmentation and named entity
  recognition}.
\newblock In \emph{Proceedings of the Fifth {SIGHAN} Workshop on {C}hinese
  Language Processing}, pages 108--117, Sydney, Australia. Association for
  Computational Linguistics.

\bibitem[{Lewis et~al.(2019)Lewis, Liu, Goyal, Ghazvininejad, Mohamed, Levy,
  Stoyanov, and Zettlemoyer}]{lewis2019bart}
Mike Lewis, Yinhan Liu, Naman Goyal, Marjan Ghazvininejad, Abdelrahman Mohamed,
  Omer Levy, Ves Stoyanov, and Luke Zettlemoyer. 2019.
\newblock Bart: Denoising sequence-to-sequence pre-training for natural
  language generation, translation, and comprehension.
\newblock \emph{arXiv preprint arXiv:1910.13461}.

\bibitem[{Li et~al.(2022)Li, Dai, Tang, Feng, Zhou, Qiu, Xu, and
  Shi}]{li2022markbert}
Linyang Li, Yong Dai, Duyu Tang, Zhangyin Feng, Cong Zhou, Xipeng Qiu, Zenglin
  Xu, and Shuming Shi. 2022.
\newblock Markbert: Marking word boundaries improves chinese bert.
\newblock \emph{arXiv preprint arXiv:2203.06378}.

\bibitem[{Liang et~al.(2020)Liang, Yu, Jiang, Er, Wang, Zhao, and
  Zhang}]{bond2020}
Chen Liang, Yue Yu, Haoming Jiang, Siawpeng Er, Ruijia Wang, Tuo Zhao, and Chao
  Zhang. 2020.
\newblock Bond: Bert-assisted open-domain named entity recognition with distant
  supervision.
\newblock In \emph{Proc. of SIGKDD}.

\bibitem[{Liu et~al.(2021)Liu, Fu, Zhang, and Xiao}]{liu-etal-2021-lexicon}
Wei Liu, Xiyan Fu, Yue Zhang, and Wenming Xiao. 2021.
\newblock \href {https://doi.org/10.18653/v1/2021.acl-long.454} {Lexicon
  enhanced {C}hinese sequence labeling using {BERT} adapter}.
\newblock In \emph{Proceedings of the 59th Annual Meeting of the Association
  for Computational Linguistics and the 11th International Joint Conference on
  Natural Language Processing (Volume 1: Long Papers)}, pages 5847--5858,
  Online. Association for Computational Linguistics.

\bibitem[{{Liu} et~al.(2019){Liu}, {Ott}, {Goyal}, {Du}, {Joshi}, {Chen},
  {Levy}, {Lewis}, {Zettlemoyer}, and {Stoyanov}}]{roberta}
Yinhan {Liu}, Myle {Ott}, Naman {Goyal}, Jingfei {Du}, Mandar {Joshi}, Danqi
  {Chen}, Omer {Levy}, Mike {Lewis}, Luke {Zettlemoyer}, and Veselin
  {Stoyanov}. 2019.
\newblock \href {http://arxiv.org/abs/1907.11692} {{RoBERTa: A Robustly
  Optimized BERT Pretraining Approach}}.
\newblock \emph{arXiv e-prints}, page arXiv:1907.11692.

\bibitem[{Ma and Hovy(2016)}]{ma-hovy-2016-end}
Xuezhe Ma and Eduard Hovy. 2016.
\newblock \href {https://doi.org/10.18653/v1/P16-1101} {End-to-end sequence
  labeling via bi-directional {LSTM}-{CNN}s-{CRF}}.
\newblock In \emph{Proceedings of the 54th Annual Meeting of the Association
  for Computational Linguistics (Volume 1: Long Papers)}, pages 1064--1074,
  Berlin, Germany. Association for Computational Linguistics.

\bibitem[{Manning et~al.(2014)Manning, Surdeanu, Bauer, Finkel, Bethard, and
  McClosky}]{corenlp}
Christopher Manning, Mihai Surdeanu, John Bauer, Jenny Finkel, Steven Bethard,
  and David McClosky. 2014.
\newblock The {S}tanford {C}ore{NLP} natural language processing toolkit.
\newblock In \emph{Proc. of ACL: System Demonstrations}.

\bibitem[{Munikar et~al.(2019)Munikar, Shakya, and Shrestha}]{munikar2019fine}
Manish Munikar, Sushil Shakya, and Aakash Shrestha. 2019.
\newblock Fine-grained sentiment classification using bert.
\newblock In \emph{2019 Artificial Intelligence for Transforming Business and
  Society (AITB)}, volume~1, pages 1--5. IEEE.

\bibitem[{OpenAI(2022)}]{chatgpt}
OpenAI. 2022.
\newblock Chatgpt: https://chat.openai.com/.

\bibitem[{Ott et~al.(2019)Ott, Edunov, Baevski, Fan, Gross, Ng, Grangier, and
  Auli}]{fairseq}
Myle Ott, Sergey Edunov, Alexei Baevski, Angela Fan, Sam Gross, Nathan Ng,
  David Grangier, and Michael Auli. 2019.
\newblock fairseq: A fast, extensible toolkit for sequence modeling.
\newblock In \emph{Proc. of NAACL-HLT: Demonstrations}.

\bibitem[{Peters et~al.(2018)Peters, Neumann, Iyyer, Gardner, Clark, Lee, and
  Zettlemoyer}]{elmo}
Matthew~E. Peters, Mark Neumann, Mohit Iyyer, Matt Gardner, Christopher Clark,
  Kenton Lee, and Luke Zettlemoyer. 2018.
\newblock Deep contextualized word representations.
\newblock In \emph{Proceedings of the 2018 Conference of the North {A}merican
  Chapter of the Association for Computational Linguistics: Human Language
  Technologies, Volume 1 (Long Papers)}.

\bibitem[{Radford et~al.(2018)Radford, Narasimhan, Salimans, Sutskever
  et~al.}]{radford2018improving}
Alec Radford, Karthik Narasimhan, Tim Salimans, Ilya Sutskever, et~al. 2018.
\newblock Improving language understanding by generative pre-training.

\bibitem[{Raffel et~al.(2020)Raffel, Shazeer, Roberts, Lee, Narang, Matena,
  Zhou, Li, and Liu}]{raffel2020exploring}
Colin Raffel, Noam Shazeer, Adam Roberts, Katherine Lee, Sharan Narang, Michael
  Matena, Yanqi Zhou, Wei Li, and Peter~J Liu. 2020.
\newblock Exploring the limits of transfer learning with a unified text-to-text
  transformer.
\newblock \emph{The Journal of Machine Learning Research}, 21(1):5485--5551.

\bibitem[{Shelmanov et~al.(2021)Shelmanov, Puzyrev, Kupriyanova, Belyakov,
  Larionov, Khromov, Kozlova, Artemova, Dylov, and
  Panchenko}]{electra-res-active}
Artem Shelmanov, Dmitri Puzyrev, Lyubov Kupriyanova, Denis Belyakov, Daniil
  Larionov, Nikita Khromov, Olga Kozlova, Ekaterina Artemova, Dmitry Dylov, and
  Alexander Panchenko. 2021.
\newblock \href {https://doi.org/10.18653/v1/2021.eacl-main.145} {Active
  learning for sequence tagging with deep pre-trained models and bayesian
  uncertainty estimates}.
\newblock pages 1698--1712.

\bibitem[{Socher et~al.(2013)Socher, Perelygin, Wu, Chuang, Manning, Ng, and
  Potts}]{sst5}
Richard Socher, Alex Perelygin, Jean Wu, Jason Chuang, Christopher~D. Manning,
  Andrew Ng, and Christopher Potts. 2013.
\newblock Recursive deep models for semantic compositionality over a sentiment
  treebank.
\newblock In \emph{Proc. of EMNLP}.

\bibitem[{Stanislawek et~al.(2019)Stanislawek, Wr{\'o}blewska, W{\'o}jcicka,
  Ziembicki, and Biecek}]{stanislawek-etal-2019-named}
Tomasz Stanislawek, Anna Wr{\'o}blewska, Alicja W{\'o}jcicka, Daniel Ziembicki,
  and Przemyslaw Biecek. 2019.
\newblock \href {https://doi.org/10.18653/v1/K19-1058} {Named entity
  recognition - is there a glass ceiling?}
\newblock In \emph{Proceedings of the 23rd Conference on Computational Natural
  Language Learning (CoNLL)}, pages 624--633, Hong Kong, China. Association for
  Computational Linguistics.

\bibitem[{Sun et~al.(2020)Sun, Fan, Han, Sun, Meng, Wu, and Li}]{sun2020self}
Zijun Sun, Chun Fan, Qinghong Han, Xiaofei Sun, Yuxian Meng, Fei Wu, and Jiwei
  Li. 2020.
\newblock Self-explaining structures improve nlp models.
\newblock \emph{arXiv preprint arXiv:2012.01786}.

\bibitem[{Tan and Zhang(2008)}]{TAN20082622}
Songbo Tan and Jin Zhang. 2008.
\newblock An empirical study of sentiment analysis for chinese documents.
\newblock \emph{Expert Systems with Applications}, 34(4):2622--2629.

\bibitem[{Tjong Kim~Sang and De~Meulder(2003)}]{conll2003ner}
Erik~F. Tjong Kim~Sang and Fien De~Meulder. 2003.
\newblock Introduction to the {C}o{NLL}-2003 shared task: Language-independent
  named entity recognition.
\newblock In \emph{Proc. of NAACL}.

\bibitem[{Touvron et~al.(2023{\natexlab{a}})Touvron, Lavril, Izacard, Martinet,
  Lachaux, Lacroix, Rozi{\`e}re, Goyal, Hambro, Azhar et~al.}]{llama1}
Hugo Touvron, Thibaut Lavril, Gautier Izacard, Xavier Martinet, Marie-Anne
  Lachaux, Timoth{\'e}e Lacroix, Baptiste Rozi{\`e}re, Naman Goyal, Eric
  Hambro, Faisal Azhar, et~al. 2023{\natexlab{a}}.
\newblock Llama: Open and efficient foundation language models.
\newblock \emph{arXiv preprint arXiv:2302.13971}.

\bibitem[{Touvron et~al.(2023{\natexlab{b}})Touvron, Martin, Stone, Albert,
  Almahairi, Babaei, Bashlykov, Batra, Bhargava, Bhosale et~al.}]{llama2}
Hugo Touvron, Louis Martin, Kevin Stone, Peter Albert, Amjad Almahairi, Yasmine
  Babaei, Nikolay Bashlykov, Soumya Batra, Prajjwal Bhargava, Shruti Bhosale,
  et~al. 2023{\natexlab{b}}.
\newblock Llama 2: Open foundation and fine-tuned chat models.
\newblock \emph{arXiv preprint arXiv:2307.09288}.

\bibitem[{Wan et~al.(2021)Wan, Yang, Marinov, Calliess, Zohren, and
  Dong}]{wan2021sentiment}
Xingchen Wan, Jie Yang, Slavi Marinov, Jan-Peter Calliess, Stefan Zohren, and
  Xiaowen Dong. 2021.
\newblock Sentiment correlation in financial news networks and associated
  market movements.
\newblock \emph{Scientific reports}, 11(1):3062.

\bibitem[{Wang et~al.(2018)Wang, Singh, Michael, Hill, Levy, and
  Bowman}]{wang2018glue}
Alex Wang, Amanpreet Singh, Julian Michael, Felix Hill, Omer Levy, and Samuel~R
  Bowman. 2018.
\newblock Glue: A multi-task benchmark and analysis platform for natural
  language understanding.
\newblock \emph{arXiv preprint arXiv:1804.07461}.

\bibitem[{Wang et~al.(2022{\natexlab{a}})Wang, Qiu, Zhang, Liu, Li, Wang, Wang,
  Huang, and Lin}]{easynlp}
Chengyu Wang, Minghui Qiu, Taolin Zhang, Tingting Liu, Lei Li, Jianing Wang,
  Ming Wang, Jun Huang, and Wei Lin. 2022{\natexlab{a}}.
\newblock \href {https://doi.org/10.18653/v1/2022.emnlp-demos.3} {{E}asy{NLP}:
  A comprehensive and easy-to-use toolkit for natural language processing}.
\newblock In \emph{Proc. of EMNLP: System Demonstrations}, pages 22--29, Abu
  Dhabi, UAE. Association for Computational Linguistics.

\bibitem[{Wang et~al.(2022{\natexlab{b}})Wang, Li, Meng, Zhang, Ouyang, Li, and
  Wang}]{wang2022k}
Shuhe Wang, Xiaoya Li, Yuxian Meng, Tianwei Zhang, Rongbin Ouyang, Jiwei Li,
  and Guoyin Wang. 2022{\natexlab{b}}.
\newblock $k$nn-ner: Named entity recognition with nearest neighbor search.
\newblock \emph{arXiv preprint arXiv:2203.17103}.

\bibitem[{Wolf et~al.(2020)Wolf, Debut, Sanh, Chaumond, Delangue, Moi, Cistac,
  Rault, Louf, Funtowicz, Davison, Shleifer, von Platen, Ma, Jernite, Plu, Xu,
  Scao, Gugger, Drame, Lhoest, and Rush}]{transformers}
Thomas Wolf, Lysandre Debut, Victor Sanh, Julien Chaumond, Clement Delangue,
  Anthony Moi, Pierric Cistac, Tim Rault, Rémi Louf, Morgan Funtowicz, Joe
  Davison, Sam Shleifer, Patrick von Platen, Clara Ma, Yacine Jernite, Julien
  Plu, Canwen Xu, Teven~Le Scao, Sylvain Gugger, Mariama Drame, Quentin Lhoest,
  and Alexander~M. Rush. 2020.
\newblock Transformers: State-of-the-art natural language processing.
\newblock In \emph{Proc. of EMNLP: System Demonstrations}.

\bibitem[{Xia et~al.(2022)Xia, Artetxe, Du, Chen, and
  Stoyanov}]{xia2022prompting}
Mengzhou Xia, Mikel Artetxe, Jingfei Du, Danqi Chen, and Ves Stoyanov. 2022.
\newblock Prompting electra: Few-shot learning with discriminative pre-trained
  models.
\newblock \emph{arXiv preprint arXiv:2205.15223}.

\bibitem[{Xin et~al.(2020)Xin, Tang, Lee, Yu, and Lin}]{xin-etal-2020-deebert}
Ji~Xin, Raphael Tang, Jaejun Lee, Yaoliang Yu, and Jimmy Lin. 2020.
\newblock \href {https://doi.org/10.18653/v1/2020.acl-main.204} {{D}ee{BERT}:
  Dynamic early exiting for accelerating {BERT} inference}.
\newblock In \emph{Proceedings of the 58th Annual Meeting of the Association
  for Computational Linguistics}, pages 2246--2251, Online. Association for
  Computational Linguistics.

\bibitem[{Xu et~al.(2023)Xu, Xu, Wang, Liu, Zhu, and McAuley}]{xu2023small}
Canwen Xu, Yichong Xu, Shuohang Wang, Yang Liu, Chenguang Zhu, and Julian
  McAuley. 2023.
\newblock Small models are valuable plug-ins for large language models.
\newblock \emph{arXiv preprint arXiv:2305.08848}.

\bibitem[{Yang et~al.(2018)Yang, Liang, and Zhang}]{yang-etal-2018-design}
Jie Yang, Shuailong Liang, and Yue Zhang. 2018.
\newblock \href {https://aclanthology.org/C18-1327} {Design challenges and
  misconceptions in neural sequence labeling}.
\newblock In \emph{Proceedings of the 27th International Conference on
  Computational Linguistics}, pages 3879--3889, Santa Fe, New Mexico, USA.
  Association for Computational Linguistics.

\bibitem[{Yang et~al.(2020)Yang, Wang, Phadke, Wickner, Mancini, Blumenthal,
  and Zhou}]{yang2020development}
Jie Yang, Liqin Wang, Neelam~A Phadke, Paige~G Wickner, Christian~M Mancini,
  Kimberly~G Blumenthal, and Li~Zhou. 2020.
\newblock Development and validation of a deep learning model for detection of
  allergic reactions using safety event reports across hospitals.
\newblock \emph{JAMA Network Open}, 3(11):e2022836--e2022836.

\bibitem[{Yang and Zhang(2018)}]{ncrfpp}
Jie Yang and Yue Zhang. 2018.
\newblock {NCRF}++: An open-source neural sequence labeling toolkit.
\newblock In \emph{Proc. of ACL, System Demonstrations}.

\bibitem[{Yang et~al.(2017)Yang, Zhang, and Dong}]{yang2017neural}
Jie Yang, Yue Zhang, and Fei Dong. 2017.
\newblock Neural reranking for named entity recognition.
\newblock In \emph{Proceedings of the International Conference Recent Advances
  in Natural Language Processing, RANLP 2017}, pages 784--792.

\bibitem[{Zacharias et~al.(2018)Zacharias, Barz, and
  Sonntag}]{zacharias2018survey}
Jan Zacharias, Michael Barz, and Daniel Sonntag. 2018.
\newblock A survey on deep learning toolkits and libraries for intelligent user
  interfaces.
\newblock \emph{arXiv preprint arXiv:1803.04818}.

\bibitem[{Zaheer et~al.(2020)Zaheer, Guruganesh, Dubey, Ainslie, Alberti,
  Ontanon, Pham, Ravula, Wang, Yang et~al.}]{zaheer2020big}
Manzil Zaheer, Guru Guruganesh, Kumar~Avinava Dubey, Joshua Ainslie, Chris
  Alberti, Santiago Ontanon, Philip Pham, Anirudh Ravula, Qifan Wang, Li~Yang,
  et~al. 2020.
\newblock Big bird: Transformers for longer sequences.
\newblock \emph{Advances in NeurIPS}.

\bibitem[{Zhang et~al.(2020)Zhang, Wang, and Yang}]{zhang2020lattice}
Yue Zhang, Yile Wang, and Jie Yang. 2020.
\newblock Lattice lstm for chinese sentence representation.
\newblock \emph{IEEE/ACM Transactions on Audio, Speech, and Language
  Processing}, 28:1506--1519.

\end{thebibliography}



\end{document}